# Ensuring Dataset Quality for Machine Learning Certification

S. Picard, C. Chapdelaine
Safran and IRT Saint-Exupéry
Magny-les-Hameaux, France
Sylvaine.picard@safrangroup.com
camille.chapdelaine@safrangroup.com

C. Cappi, L. Gardes
SNCF
Saint Denis, France
cyril.cappi@sncf.fr
l.gardes@sncf.fr

E. Jenn
Thales AVS and IRT Saint-Exupéry
Toulouse, France
eric.jenn@irt-saintexupery.com

B. Lefevre
Thales
Mérignac, France
baptiste.lefevre@fr.thalesgroup.com

T. Soumarmon
Continental and IRT Saint-Exupéry
Toulouse, France
thomas.soumarmon@continental-corporation.com

*Abstract* — In this paper, we address the problem of dataset quality in the context of Machine Learning (ML)-based critical systems. We briefly analyse the applicability of some existing standards dealing with data and show that the specificities of the ML context are neither properly captured nor taken into account. As a first answer to this concerning situation, we propose a dataset specification and verification process, and apply it on a signal recognition system from the railway domain. In addition, we also give a list of recommendations for the collection and management of datasets. This work is one step towards the dataset engineering process that will be required for ML to be used on safety critical systems.

## I. INTRODUCTION

Datasets are the main inputs for machine learning (ML) algorithms. Hence, compliance of an ML system to its *intended function* obviously strongly depends on the *quality* of those data.

*Quality* is defined by the ISO9000 as "the degree to which a set of inherent characteristics of an object fulfils requirements" [1, Sec. 3.6.2] or, according to the DO-200B [2], "[the] degree or level of confidence that the data provided meets the requirements of the user".

In this paper, we make explicit those "inherent characteristics", explain their relations to the system requirements, and show how to verify them on a train signal recognition system application.

Ensuring the quality of a ML system dataset is difficult, in particular because the relation between the characteristics of the data and their effect on the compliance of the ML system to its requirements is notoriously complex and difficult to establish. Even worse, the set of characteristics itself is difficult to establish, as we will illustrate later on our use case.

Various sets of data quality characteristics or *dimensions* have been proposed in standards or in the literature [2],[3]. Among them, we can mention: *Accuracy, Accessibility, Consistency, Relevance and Fitness, Timeliness, Traceability,* *Usability*. Some quality dimensions can be universal like *Accuracy* or context specific like *Timeliness*.

These quality dimensions (and the associated quality management activities) have initially been developed in the context of statistical studies or forecasting. ML brings a whole new set of difficulties since those data are not only used to *extract* information but to build a model that will be used to process other data and determine the behaviour of a system, and more specifically, the behaviour of a critical system submitted to a certification process.

The objective of certification is to demonstrate that a system satisfies some high-level objectives, such as, in the aeronautical domain, those stated in the CS 25.1309 [4]. Demonstration of compliance usually rely on detailed industry standard recognized by the certification authorities as acceptable means of compliance. However, existing means of compliance (e.g., ED-12/DO178C [5], ED-79/ARP4754 [6], etc.) have been designed for "classical" systems based on explicit, deterministic implementations. Those means of compliance need to be complemented to address ML methods based on statistical reasoning. In addition, classical methods usually place the emphasis on the processing of data (by software or hardware), not on the data themselves. Generally, data are inputs *processed by* the system (acquired via sensors), parameters of the system (e.g., configuration parameters, calibration data, etc.), or databases used by the system (e.g., navigation databases, obstacle databases, etc.). Those data may determine more or less strongly the behaviour of the system, but they do not describe the logic of this behaviour, as it is the case for machine learning. To some extent, data in ML play a similar role as binary code in classical systems. Therefore, existing means of compliance need to be revised to address this new role of data, possibly relying on existing practices (see Section III).

Therefore, we propose in this paper a workflow (i.e., activities and artefacts) to address the question of dataset quality in the context of the development of a ML-based certified system. This workflow is organized around the following main artefacts:
- A Dataset Definition Standard (DDS),
- A Dataset Requirement Specification (DRS)

- A Dataset Verification Plan (DVP)

In the following, we describe those artefacts and the overall workflow that produces and consumes them. We illustrate them in our concrete use case.

The rest of the paper is organized as follows: Section II gives an overview of related works in the general context and in the context of machine learning; Section III presents the existing practices about data quality in the aeronautical and railway domains, and analyse how these practices relate to the problem of data quality for ML; Section IV briefly introduces our use case; Section V presents our workflow and illustrates its application in our use case; finally, section 0 concludes the paper.

## II. RELATED WORKS

Many authors have discussed data quality dimensions [3], [7]–[11]. A comprehensive presentation of data quality is proposed in [3], from which the definitions of the following data quality dimensions are taken:

- Accuracy: a measure of the correctness of the content of the data (which requires an authoritative source of reference to be identified and accessible)
- Accessibility: refers to how easily data can be accessed; the awareness of data users of what data is being collected and knowing where it is located
- Consistency: determines the extent to which distinct data instances provide non-conflicting information about the same underlying data object
- Timeliness: data are up to date (current) and information is available on time
- Traceability: the lineage of the data is verifiable,
- Usability: refers to the extent to which data can be accessed and understood
- Relevance: refers to the extent to which the data meet the needs of users. Information needs may change and is important that reviews take place to ensure data collected is still relevant for decision makers.

These data quality dimensions can be used in different contexts such as statistical analysis or software system certification. For this reason, their importance and impact and even definition can vary depending on the domain of application. An interesting study of the relation between data and safety in the context of automotive application is found in [12]. The authors explore the impact of data quality in a critical system. In this work, data is classically a software input. In the context of ML, particularly with Deep Learning (DL), new difficulties arise:

- Training datasets may be part of the system specification, or possibly, the specification itself
- Small biases on data can have nearly unpredictable consequences on the system behaviour
- The overall system performance is assessed, at least primarily, on datasets.

Several recent works focus on the question of data qualities for Machine Learning, including [13] which proposes a survey of the literature complemented by a series of interview with data scientists concerning data quality models for Machine Learning, or [14] which propose a framework to manage the data quality lifecycle in this context. Our work exploits those results by proposing a curated list of recommendations (the DDS, see Section V.B) that could form the basis for a future standard dedicated to dataset management for safety critical ML systems.

## III. STANDARDS AND DATASETS

Developing standards and regulations for ML-based certified systems is an on-going effort. In particular, there is currently no agreed standard and no regulation on datasets applicable to such systems. However, in a different but similar context, datasets have been used for a long time in a certified environment. This is for instance the case of the navigation data used to guide aircrafts. Hereafter, we present the standard that is applicable to such data.

### A. Data quality in the aeronautical domain: the ED-76A/DO-200B

In aviation, aeronautical data are managed according to a process defined in ED-76A/DO-200B [2] entitled 'Standards for Processing Aeronautical Data'.

The ED-76A/DO-200B is applicable to aeronautic databases used by ground or airborne systems that may have an effect on safety. A typical example is the Navigation System Data Base[1] [15] used by Flight Management Systems. It shall be noted that compliance to the DO-200B does not say anything about the data themselves; it only provides "guidance and requirements for data processing activities that may occur between data origination and end use" [2].

This standard is recognized as an acceptable means of compliance with aviation safety regulations for providers of data services. Even if this standard is not directly applicable to Machine Learning, we propose here a short description of the proposed process, in order to explain how data are usually managed in a certified environment.

According to ED-76A/DO-200B, the compliance of data with the requirements is demonstrated based on:

- A definition of Data Quality Requirements (DQRs),
- A definition of Data Processing Requirements (DPRs),
- A definition of Quality Management Requirements (DMRs).

DATA QUALITY REQUIREMENTS (DQRs)

In the context of the ED-76A/DO-200B, the data quality requirements cover the following qualities: accuracy, resolution, assurance level (confidence that the data is not corrupted), traceability, timeliness, completeness, and format.

DATA PROCESSING REQUIREMENTS (DPRs)

---

[1] The "ARINC 424" navigation database.

In order to ensure that data are processed in a way that preserves the DQRs, data processing requirements are also defined. These requirements include the following aspects:
- Definition of the data processing procedure
- Data configuration management
- Competency management
- Tool qualification
- Cybersecurity

QUALITY MANAGEMENT REQUIREMENTS (QMRs)

Quality management requirements ensure that data meets the agreed DQRs and that the relevant data processing procedures are indeed applied. These requirements define how DQRs and compliance with data processing procedures are checked (using quality reviews, audits and controls for example).

*B. Data quality in the railway domain: the EN50128*

For railway systems, a relevant standard is the EN50128 [16]. The latest version of this standard does not really cover the case of the introduction of ML techniques in railway systems. It simply and briefly mentions IA techniques as a possible means to forecast and correct defaults, and determine maintenance action [16, Sec. D.1]. In addition, this only concerns SIL0 systems (i.e., system with the less demanding Safety Integrity level).

The case for the data used to configure a software application is addressed in [16, Sec. 8], "Application data development". The objective of the data development process is (i) to obtain the data correctly from the concerned installation and the verification of the behaviour of the system, and (ii) to evaluate the development process used to produce those data. This process is based on a preparation plan that ensures that the application data are "complete, correct and compatible with each other and with the application". Techniques for preparing the data are recommended depending on the level of SIL [16, Sec. A], Table A.11, but the recommendations remain very general. Nevertheless, it is worth noting that formal proof is *highly recommended* for the verification of data used in SIL3 and 4 systems.

[17] gives an interesting example on how such formal verification can be performed on configuration data representing the geographical arrangement of track equipment, but it is worth noting that – in this case – data show a strong level of organization that facilitate the mathematical expression, and then verification, of properties (for instance, the graph of "areas" must be strongly connected).

*C. Other domains*

A comprehensive analysis of all standards being out of the scope of this paper, we will simply mention the SOTIF [18], a standard focused on some of the concerns raised by the use of ML for road vehicles. This standard proposes a series of measures to address hazardous behaviours resulting from "limitations of the implemented functions", including limitations due to "the inability of [a] function to correctly comprehend the situation to operate safely" such as the one that can be encountered for "functions that use machine algorithms". However, the standard remains at system level, and does not provide any specific recommendation about the data collection process, besides stating that "the data collection and learning system to be developed according to safety standards, with attention given to reducing hazards such as unintended bias or distortion in the collected data" [18, Appendix G].

*D. Applicability to ML datasets*

The general approach proposed in the ED-76A/DO-200B or EN50128 standards is *overall* relevant and applicable to the datasets of a ML-based certified product. In particular, defining data quality, data processing, quality management requirements, and verification procedures are necessary in order to ensure that the datasets actually have the expected quality attributes.

However, it shall be noticed that the datasets considered in both the ED-76A/DO-200B and EN50128 also show significant differences with respect to the datasets used in ML, in particular:
- The datasets concerned by the ED-76A/DO-200B or EN50128 are *inputs* of some data processing activity (e.g., computation of an aircraft trajectory, management of the state of some track side signalling. Even if those data may strongly determine the actual behaviour of the system, the set of possible behaviours is still described by some software or hardware description code. In practice, some properties on the system may be verified by the sole analysis of the code, without even considering the actual data. In the case of ML, the actual behaviour is largely determined by the data used to train the model.
- The set of requirements with which the datasets concerned by the ED-76A/DO-200B or
EN5128 must comply can be fairly easily expressed. Indeed, it derives from the requirements applicable to the function. For instance, the precision of the geographical locations of runways and taxiways on an airport map is completely determined by the required precision of the trajectory and the algorithms used to compute this trajectory.

  In the case of the dataset used by ML, this link also exists, but is much more difficult to establish formally since the "algorithm" itself derives from the data.
- The completeness of the databases considered in ED-76A/DO-200B or EN5128 is generally easy to define and verify because the completeness criterion can be expressed in simple and clear terms such as "all runways and taxiways of all airport in France", "all track signals in this areas", etc. Things are much more complicated when considering ML datasets since the input space is usually infinite, and strict completeness must be replaced by some statistical criterion (sampling criterion).
- Finally, in ML, the dataset may even be considered as part of the specification (or even the complete specification) of the system in the specific cases where the behaviour can only be specified by means of a series of (input, output) pairs.

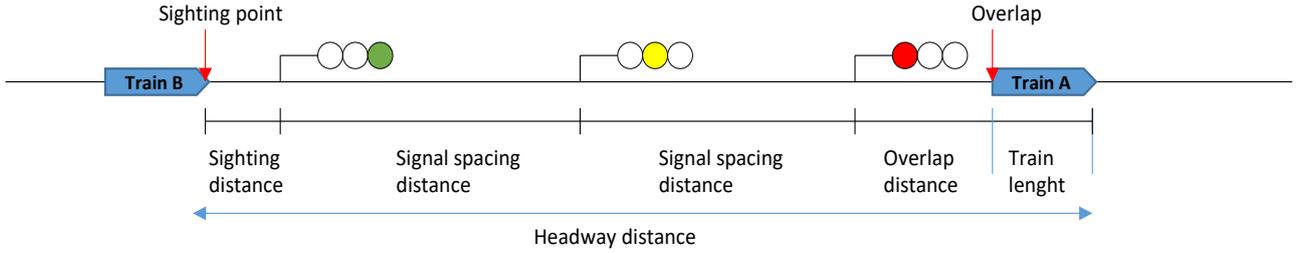

Figure 1. Signalling principles

## IV. THE RAILWAY USE CASE

To illustrate our approach, we have considered an Automatic Signal Processing system (ASP). An ASP is aimed at recognizing the state of a light signal applicable to the train, a task that is currently performed by train drivers. The reader can find in Harb *et al*. [19] a large public dataset illustrating this use case.

The ASP shall remain vision-based in order to limit the impact on the infrastructure, and limit the cost of its deployment. In our study, the ASP uses ML-based algorithm.

This use case shows two interesting properties with respect to the objective of our study: it implements a fairly simple task (recognizing a light signal) *and* it operates in an open and weakly structured environment.

Before describing the system architecture, an overview of the principles of railway signalling is proposed.

### A. Railway Signalling Principles

In the railway field, signalling establishes a link between some infrastructure elements (track, bridges, etc.) and the rolling stock (trains) [20]. Signalling has a dual role. On the one hand, it plays a significant role on safety by authorizing/forbidding/stopping the movements of trains, protecting the passengers and workers, etc. On the other hand, it allows the efficient operation of traffic by regulating trains. It actually is one of the most important component for the railway system [21]–[23]. This system is considered to be SIL 4 which is roughly equivalent to a DAL A in the aeronautical domain.

For safety reasons, trains must be separated by a minimum interval of time, or *headway* (see Figure 1). This parameter is a key input in calculating the overall route capacity of any transit system. Thereby, a system requiring large headways causes more empty spaces than passenger capacity, which lowers the total number of passengers or cargo quantity transported through a given length of line. The role of light signals is to ensure the minimal headways that both ensures safety and maximizes the line capacity.

### B. ASP description

Figure 2 (left) describes the driver's actions when the train is approaching a signal [24]. The ASP shall perform the same tasks. For operational reasons, the ASP shall work in weather conditions and lines contexts that are both very complex and variable. In addition, the signal can be occluded or damaged. Figure 2 (right) gives an overview of the system architecture. The machine-learning algorithm under consideration is depicted by the black box.

The input of the algorithm is one RGB image and one box that bounds the signalling equipment to process. For each image, the system shall produce an estimation of the signal state. The ML algorithm is trained from images captured by a camera installed on a real equipment.

## V. OUR PROPOSAL

Considering that existing standards such as the ED-76A/DO-200B or the EN50128 to ML Datasets only cover part of the problem, a new approach to develop datasets is needed. In this paper, we suggest to rely on three main documents:

- The Dataset Definition Standard (DDS), which expresses general recommendations about the building of datasets

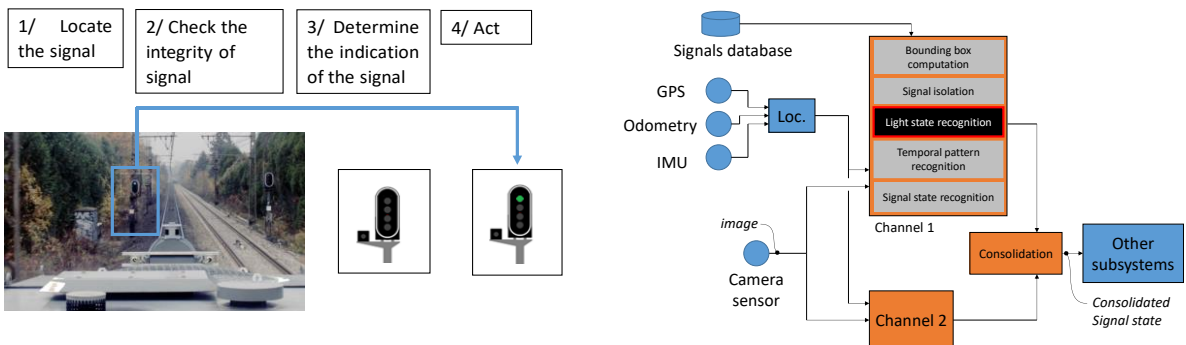

Figure 2. Overview of the ASP function (left) and Architecture (right).
Channel 1 and channel 2 are redundant systems required for safety reason

- The Dataset Requirement Specification (DRS), which collects all requirements applicable to the dataset
- The Dataset Verification Plan (DVP), which defines procedures to verify the compliance of the dataset with its specification.

The content of these documents is respectively detailed in Sections B, C, and D below. Those documents are developed through a tight collaboration between various actors.

*A. Actors and responsibilities*

The construction and verification of the dataset requires the collaboration of three different experts: the *application expert*, the *acquisition systems expert,* and the *machine learning expert*.

The application expert brings his knowledge of the use case at hand: operational scenarios, consequences of a system behaviour, operational conditions, certification standards, and usage of the field. Above all, he ensures the dataset representativeness. He has a fundamental role in the DRS production. The second expert is the *Acquisition System* expert. It is common to use "off-the-shelf" data or acquisition system to create a dataset, but it is a bad practice. The acquisition expert shall bring his knowledge about important parameters, equivalence classes, potential perturbations and their impact on the data, correct handling of the data (e.g. compression impact). He can anticipate consequences of the operational constraints. His role is very important in the DRS and DVP creation.

Finally, the third expert is the *Machine Learning* expert. Machine learning expert's role is to ensure that good practices in datasets creation are fully applied (e.g. no mix of data between training and testing which can easily occur in case of images coming from video acquisitions). For instance, he guarantees the statistical quality of the dataset: correct bias handling, dataset size, coherence of the annotation, etc. He brings the knowledge of the DDS requirements and can work to its adaptation to the application during the DRS construction.

*B. Dataset Definition Standard (DDS)*

The *Dataset Definition Standard* [25] collects generic recommendations stemming from the machine learning field. All of those recommendations have a positive effect on the quality of the dataset, and the relevant artefacts and procedures in the development of a critical system shall comply with all these recommendations.

In the following, we present some elements of our DDS. However, defining such a standard is a significant and difficult work which should ideally be carried out jointly by the machine learning community and by standardization bodies. DDS extends the classes of requirements given by ED-76A/DO-200B about data quality [26, Sec. 2.3] and data processing procedures [26, Sec. 2.4] to take into account the specificities of Machine Learning.

Requirements on data qualities are extended to cover issues such as
- Representativeness and biases
- Reliability (e.g., absence of outliers)
- Redundancy and consistency
- Etc.

The set of procedures are also extended to address
- The data collection process (e.g., the size of the dataset w.r.t the complexity of the model)
- The annotation process, in order to ensure an adequate labelling of data in supervised learning
- The data protection process, in order to prevent the unexpected alteration of data during the data pre-processing phases
- The constitution of the learning, validation, and test datasets, in order to maintain the quality of the performance assessment
- Etc.

The DDS contains generic requirements applicable to the development of any ML-based system. Genericity is achieved by abstraction, i.e., statement of general rules (e.g., splitting of datasets) or by exhaustion (coverage of all types of data that can be encountered in embedded systems). To be applied in a specific use case. This is done in the DRS (see next section).

The document is organized according to the data quality attributes. For each of them, we
- *define* the attribute:

"**[DEF-21.1]** [...] A sample is said to be *representative* if it contains key features with the same distribution of the actual population […]"

- *express* the objectives and *explain* why it is relevant from a safety perspective :

"**[OBJ-21.1]** […] The dataset used for the system development shall be representative of the situations that will be encountered by the system in operation. Representativeness is required for the learning phase to capture correctly the expected behaviour of the system in operation. It is also required for the performance evaluation of the system to be significant. […], "

- *state* recommendations:

"**[REC-21-1]** Acquisition of the data used to build the dataset must be done with an acquisition chain as close as possible to the one that will be used in operation. Any difference shall be justified and its effects on the learning process must be assessed. […]"

"**[REC-21-2]** [...] Dataset is composed of independent and identically distributed data (i.i.d) […]"

"**[REC-21-3]** The different classes that a system must discriminate must be represented in the dataset in the same proportion as in operations. Any difference must be justified (*) […]"

(*) Some biases may be justified to orient the system towards safety, or for fine grain performance analysis.

"**[REC-21-4]** The dataset shall be large enough to allow performance estimations with the appropriate confidence level. The size of the dataset may be estimated according to bounds derived from Bernstein Inequalities (see e.g., [27])."

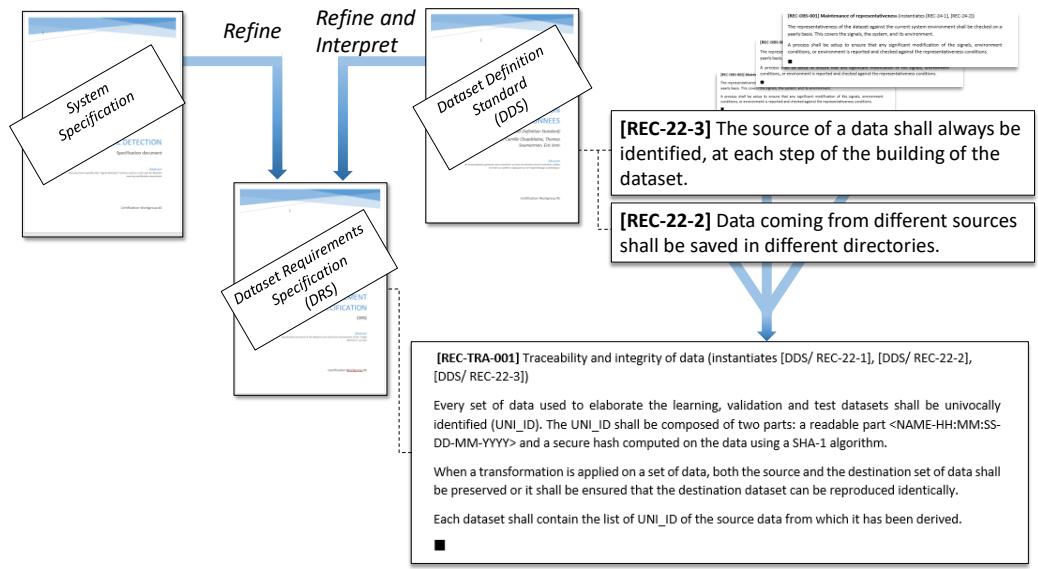

Figure 3. Relation between DDS and DRS

Note that those recommendations sometimes require to be interpreted in the specific context of the system under design. For instance, let's take the following recommendation:
"**[REC-42-2]** Redundancy shall be avoided between the validation and test datasets. […]".
In the context of the signal recognition system, the input is a series of images captured in sequence, but our algorithm processes its inputs "image by image", as if they were independent, uncorrelated. So, when splitting into train, validation, and test subsets, care shall be taken not to spread quasi identical images in the train and test subsets. If this recommendation is not applied, there is a risk for the tests to give over-optimistic performance assessment.

The first version of the document contains around 50 recommendations. Simple examples of the kind of rules found in the DDS are shown in Figure 3.

*C. Dataset Requirement Specification (DRS)*

The Dataset Requirement Specification (DRS) captures and refines the system requirements allocated to the dataset component of the system. The dataset specification also contains *derived requirements*, i.e. requirements that are not directly traceable to the system specification. This is the case of the DRS requirements deriving from the DDS (see Figure 3).
Requirements assigned to the dataset must satisfy the usual properties of requirements, including *validity*, *completeness* (and representativeness), and *innocuity*.
**Validity**
To some extent, the dataset can be considered as an explicit, or enumerated, version of the high-level specification of the intended function (or of part of it). So, the dataset shall contain a set of (input, output) couples that fully comply with the specification. In practice, this implies that all (input, output) pair must be validated with respect to the specification before being used in the training process.
**Completeness and representativeness**
The specification shall also be *complete*. In the case of a dataset, this boils down to being representative of all foreseeable conditions (in particular, environmental conditions) that will be encountered by the system during its operational life. In our use case, for instance, it is explicitly required that the system will operate *day and night*. This is usually interpreted as "in any situation where the light level is in the range [X,Y] lumen". The feature is the "light level". (Note that this could also be interpreted as a range of elevations of the sun). So, "light level" must be one of the features to be specified in the DRS.
But most features are not mentioned in the high-level specification. For instance, it is likely that the "phase of the moon" is not considered.
So, the problem consists to define a systematic and (economically) reasonable process to identify the *relevant features* to be selected.
The relevance criterion is hard to express because it basically depends on the effect of the feature on the Machine Learning algorithm. At the time the dataset is established, the algorithm may be neither chosen nor implemented yet, so it may be very difficult to estimate a priori if a given feature will play a role on the capability of the system to perform the intended function. Note that this means that it may be wise to refine the dataset specification conjointly to the refinement of the Machine Learning system… It is the role of the three experts described above to anticipate relevant parameters and translate them in the DRS.
The selection of features may first be based on field experience. For instance, a train driver "knows" – by experience – that the speed of the train plays a significant role on the capability to recognize a signal. Contrast changes is another "feature" that may affect the performance of a human driver. Even though "speed" or "contrast changes" may play a much less significant effect on a Machine Learning component than on a human, it seems wise and safe to consider them in the list of "relevant" features. In practice, the DRS will specify

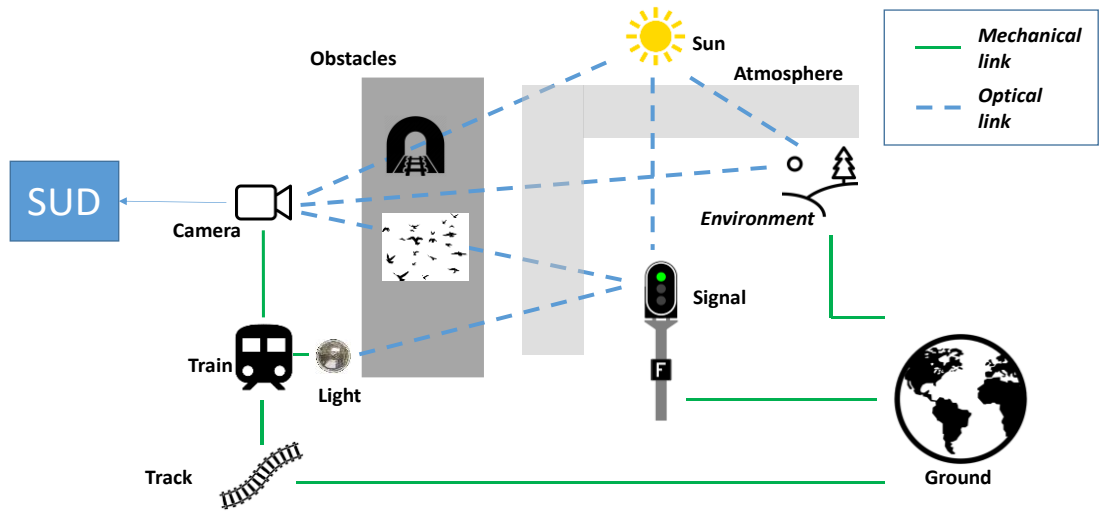

Figure 4. Building the DRS

how the train speeds and how contrast changes shall be covered by the dataset.

Conversely, other features not considered to play a role on the performance of a human (e.g. the temperature) *may* be very relevant for a Machine Learning algorithm.

So, in order to minimize the risk of missing an important feature, we definitely need a systematic approach to identify them.

The only input of our system is an image (i.e., a set of pixels), we propose to perform a systematic analysis to determine what, in the environment, may affect this image.

In practice, we have to consider the complete optical path going from the light source to the sensor, analyse the properties of all elements involved in this optical path and their relation to the image. This is illustrated on Figure 4 which shows the various components involved in the overall system, and their relations. For instance, the camera and the train are related by a mechanical link. The train and the track, the track and the ground, the signal and the ground are all related by mechanical links. Those links must be analysed carefully to determine the features to be accounted for in the definition of the dataset.

For instance, the bank angle of the track modifies the orientation of the signal on the image, and so do the vibrations of the camera, the presence of rain, etc. All those features have to be integrated in the list of relevant features.

Generally speaking, the analysis shall cover the acquisition chain but also all elements (electrical, mechanical, environmental) that may have an impact on the image.

**Innocuity**

Besides being complete, the dataset shall not specify anything else but the intended function, or if it is not the case, demonstration shall be given that any additional function has no negative effect on safety.

### D. Dataset Verification Plan (DVP)

The Dataset Verification Plan (DVP) is used to demonstrate the compliance of a particular dataset with the DRS it is supposed to implement. The DVP provides a verification procedure for each requirement of the DRS. In some cases, the procedure can be implemented by a computer program and is then described by a pseudo code (or any convenient language). In other cases, the procedure must be implemented manually and is then described as a sequence of actions.

It is not rare that datasets are created in an iterative manner, hence applying regularly the DVP to the current iterations of the dataset helps to precisely measure the compliance of the dataset with its DRS.

The DVP must be established *before* the dataset is actually created. Indeed, verification activities may require additional information besides the one that will be used for training (those metadata will be recorded with the data).

In general, automatic processing for data quality verification is preferable, as it is not impacted by human factor and can be used iteratively on the dataset (allowing to measure dataset quality improvement). It can exhaustively parse the dataset, it is in general faster and cheaper. In case of automatic verification, DVP contains the verification code or pseudo code, parsing strategy, and error tolerance.

Unfortunately, automatic verification is not always possible and some characteristics are verified manually. In this case, DVP contains a comprehensive description of the verification strategy, data sampling strategy, and decision criteria, inspectors 'skills and profiles, and verification tools.

In the following, we provide one example of procedure in the context of our use case. It illustrates an automatic verification specification. In outdoor applications, like in automatic signal state classification, the sun elevation is of great importance, because it introduces great differences in the landscape appearance (e.g. shadows) and can have high impact on images quality (e.g. flare). The sun elevation histogram can be automatically checked, for example using the Astropy[2] package. This example illustrates how constants defined by the DRS are used by the DVP. It shows also the role played by the DVP in the acquisition system specifications (e.g. need of GPS and time acquisition). After the checking procedure, the decision criteria are clearly described.

---

[2] See https://www.astropy.org.

**[REC-10-1] Sun elevation**
Language: Python
Imported Modules: astropy

Constants defined in DRS:
cDRS_SunElev_BucketTolerance = 0.1

Input data :
nDatasetNumImages : size of the dataset
aGPS_longitude: GPS latitude,
aGPS_latitude: camera GPS latitude
aImageTime : images time stamps

**Checking procedure**
```
initHistogram(aHistogramElevation)
for all dataset images:
        loc = EarthLocationFromGPS(aGPS_longitude, aGPS_latitude)
        sunElev = SunElevationFrom(loc, aImageTime)
        aHistogramElevation[sunElev] += 1

If CheckHistogramCompliant(aHistogramElevation, cDRS_SunElev_BucketTolerance)
                return REQ 101_OK
else
                return REQ 101_KO
```
**Decision criteria:**
The repartition of sun elevation in the dataset is compared to the histogram specified in the DRS. The error for each category must not exceed *10%*.

## VI. CONCLUSIONS AND FUTURE WORK

The appropriate building and management of datasets is a necessary condition to get confidence in a Machine learning based system. Learning from inadequate data (e.g. irrelevant, truncated, incomplete…) leads inevitably to an inadequate behaviour.

We have seen that the data used for the training of ML systems are by nature very different from those considered in the current quality standards applicable to safety critical systems. Their domain, structure, and relation to the system behaviour strongly differ from those of the data contained in the databases or set of configuration parameters concerned by those standards. Even though recommendations given by those standards are also applicable on ML datasets, they are definitively not sufficient.

In this paper, we have presented a first approach to engineer the dataset as an integral part the system under design. Towards that goal, we have proposed an approach and three artefacts – the Dataset Definition Standard, the Data Requirement Specification, and the Data Verification plan – used to specify and verify datasets and have applied this "approach" on an image processing use case in the railway domain.

In the next phase, we plan to analyse some of the recommendations given in the Dataset Definition Standard from a mathematical perspective (e.g., representativeness).

ACKNOWLEDGEMENTS

The authors want to thank all members of the AI Certification Workgroup – in which this work has been carried out –, the funding members of the DEEL project, the ANR, the ANITI, and the STAE RTRA for their support.